\documentclass[conference,compsoc]{IEEEtran}
\makeatletter\if@twocolumn\PassOptionsToPackage{switch}{lineno}\else\fi\makeatother

\usepackage{graphicx}
\usepackage[T1]{fontenc}
\ifCLASSOPTIONcompsoc
  \usepackage[nocompress]{cite}
\else
  \usepackage{cite}
\fi
\ifCLASSINFOpdf
\else
\fi
\usepackage{amsmath}
\usepackage[cmintegrals]{newtxmath}
\usepackage{url,multirow,morefloats,floatflt,cancel,tfrupee}
\makeatletter

\AtBeginDocument{\@ifpackageloaded{textcomp}{}{\usepackage{textcomp}}}
\makeatother
\usepackage{colortbl}
\usepackage{xcolor}
\usepackage{pifont}
\usepackage[nointegrals]{wasysym}
\makeatletter

\def\mcWidth#1{\csname TY@F#1\endcsname+\tabcolsep}

\def\cAlignHack{\rightskip\@flushglue\leftskip\@flushglue\parindent\z@\parfillskip\z@skip}
\def\rAlignHack{\rightskip\z@skip\leftskip\@flushglue \parindent\z@\parfillskip\z@skip}

\@ifundefined{etal}{}{}

\usepackage{ifxetex}
\ifxetex\else\if@twocolumn\@ifpackageloaded{stfloats}{}{\usepackage{dblfloatfix}}\fi\fi

\AtBeginDocument{
\expandafter\ifx\csname eqalign\endcsname\relax
\def\eqalign#1{\null\vcenter{\def\\{\cr}\openup\jot\m@th
  \ialign{\strut$\displaystyle{##}$\hfil&$\displaystyle{{}##}$\hfil
      \crcr#1\crcr}}\,}
\fi
}

\AtBeginDocument{%
  \@ifpackageloaded{endfloat}%
   {\renewcommand\efloat@iwrite[1]{\immediate\expandafter\protected@write\csname efloat@post#1\endcsname{}}}{\newif\ifefloat@tables}%
}%

\def\BreakURLText#1{\@tfor\brk@tempa:=#1\do{\brk@tempa\hskip0pt}}
\let\lt=<
\let\gt=>
\def\processVert{\ifmmode|\else\textbar\fi}

\@ifundefined{subparagraph}{
\def\subparagraph{\@startsection{paragraph}{5}{2\parindent}{0ex plus 0.1ex minus 0.1ex}%
{0ex}{\normalfont\small\itshape}}%
}{}

\newcommand\role[1]{\unskip}
\newcommand\aucollab[1]{\unskip}
  
\@ifundefined{tsGraphicsScaleX}{\gdef\tsGraphicsScaleX{1}}{}
\@ifundefined{tsGraphicsScaleY}{\gdef\tsGraphicsScaleY{.9}}{}
\def\checkGraphicsWidth{\ifdim\Gin@nat@width>\linewidth
	\tsGraphicsScaleX\linewidth\else\Gin@nat@width\fi}

\def\checkGraphicsHeight{\ifdim\Gin@nat@height>.9\textheight
	\tsGraphicsScaleY\textheight\else\Gin@nat@height\fi}

\def\fixFloatSize#1{}
\let\ts@includegraphics\includegraphics

\def\inlinegraphic[#1]#2{{\edef\@tempa{#1}\edef\baseline@shift{\ifx\@tempa\@empty0\else#1\fi}\edef\tempZ{\the\numexpr(\numexpr(\baseline@shift*\f@size/100))}\protect\raisebox{\tempZ pt}{\ts@includegraphics{#2}}}}

\AtBeginDocument{\def\includegraphics{\@ifnextchar[{\ts@includegraphics}{\ts@includegraphics[width=\checkGraphicsWidth,height=\checkGraphicsHeight,keepaspectratio]}}}

\DeclareMathAlphabet{\mathpzc}{OT1}{pzc}{m}{it}

\def\URL#1#2{\@ifundefined{href}{#2}{\href{#1}{#2}}}

\def\UrlOrds{\do\*\do\-\do\~\do\'\do\"\do\-}%
\g@addto@macro{\UrlBreaks}{\UrlOrds}

\edef\fntEncoding{\f@encoding}

\makeatother

\newif\ifmultipleabstract\multipleabstractfalse%
\usepackage{tabulary}
\makeatletter
\AtBeginDocument{\@ifpackageloaded{longtable}{%
\def\LT@makecaption#1#2#3{%
  \LT@mcol\LT@cols c{\hbox to\z@{\hss\parbox[t]\LTcapwidth{%
    \sbox\@tempboxa{#1{#2: } #3}%
    \ifdim\wd\@tempboxa>\hsize
      #1{#2: }\textsc{#3}%
    \else
      \hbox to\hsize{\hfil\box\@tempboxa\hfil}%
    \fi
    \endgraf\vskip\baselineskip}%
  \hss}}}
}{}}
\makeatother

   \makeatletter
  \def\fig@textbf{\textbf}
   \AtBeginDocument{\renewcommand\floatc@plain[2]{\setbox\@tempboxa\hbox{{\footnotesize#1.}\footnotesize\hskip.5em#2}%
    \ifdim\wd\@tempboxa>\hsize {\fig@textbf{\footnotesize#1.}}\footnotesize\hskip.5em#2\par
        \else\hbox to\hsize{\hfil\box\@tempboxa\hfil}\fi}}
    \makeatother
  
\usepackage[flushleft]{threeparttablex}

\usepackage{float}

\begin{document}

%


        \title{Salient Image Matting }
      
\author{
		\IEEEauthorblockN{Rahul~Deora}
    \IEEEauthorblockA{Fynd}\\[-12pt]Email: rahuldeora@fynd.com ~\\(Corresponding author)
        \vspace*{1pc}\and 
		\IEEEauthorblockN{Rishab~Sharma}
    \IEEEauthorblockA{Fynd}\\[-12pt]Email: rishabsharma@fynd.com
        \vspace*{1pc}\and 
		\IEEEauthorblockN{Dinesh~Samuel Sathia Raj}
    \IEEEauthorblockA{University of Michigan}\\[-12pt]Email: dssr@umich.edu}
  


\maketitle 

\begin{abstract}
In this paper, we propose an image matting framework called Salient Image Matting to estimate the per-pixel opacity value of the most salient foreground in an image. To deal with the large amount of semantic diversity in images, a trimap is conventionally required as it provides important guidance about object semantics to the matting process. However, creating a good trimap is often expensive and time-consuming. The SIM framework simultaneously deals with the challenge of learning on a wide range of semantics and salient object types in a fully automatic and end to end manner. Specifically, our framework is able to produce accurate alpha mattes for a wide range of foreground objects and cases where the foreground class, such as human, appears in a very different context than the train data directly from an RGB input. This is done by employing a salient object detection model to produce a trimap of the most salient object in the image in order to guide the matting model about higher level object semantics. Our framework leverages large amounts of coarse annotations coupled with a heuristic trimap generation scheme to train the trimap prediction network so it can produce trimaps for arbitrary foregrounds. Moreover, we introduce a multi-scale fusion architecture for the task of matting to better capture finer, low level opacity semantics. With high level guidance provided by the trimap network, our framework requires only a fraction of expensive matting data as compared to other automatic methods while being able to produce alpha mattes for a diverse range of inputs. We demonstrate our framework on a range of diverse images and experimental results show our framework compares favorably against state of art matting methods without the need for a trimap. 
\end{abstract}
    

%
\IEEEpeerreviewmaketitle

\section{Introduction}
Image matting refers to the problem of accurately estimating the foreground object opacity in images and video sequences. It serves as a prerequisite for a broad set of applications, such as film production, digital image editing and live streaming as it allows for realistic compositions of the foreground on novel backgrounds. Matting also plays an important role in making creative compositions of human and product images on a variety of backgrounds for creative marketing and design campaigns. Such images have diverse foregrounds and must meet high quality standards as they are usually used for commercial purposes. To easily process such images, a fully automatic matting system is desired.

\bgroup
\fixFloatSize{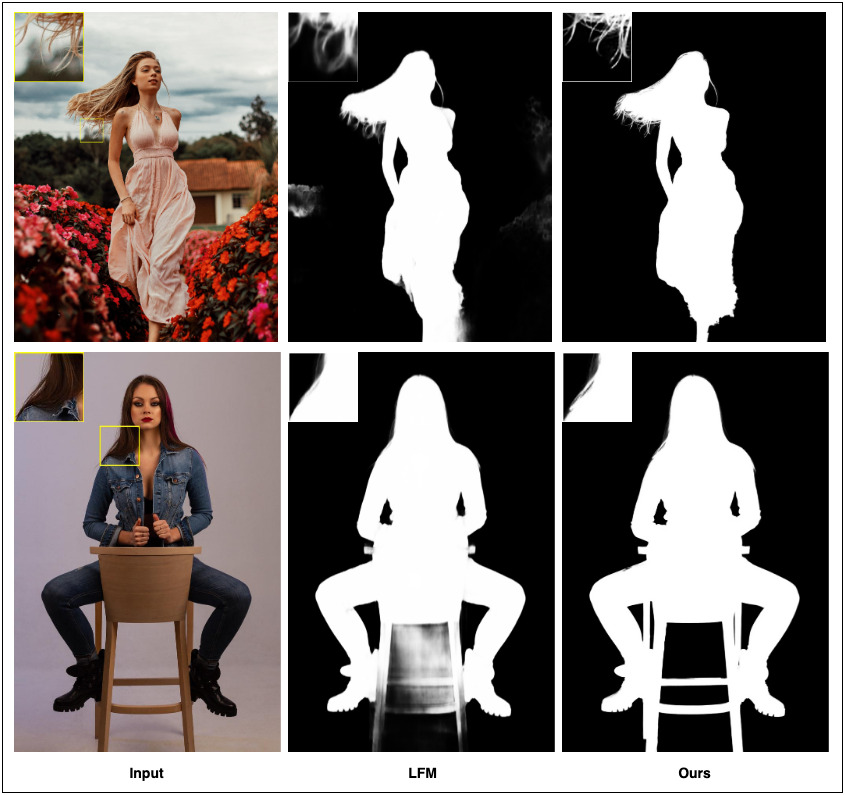}
\begin{figure}[!htbp]
\centering \makeatletter\IfFileExists{intro.png}{\includegraphics{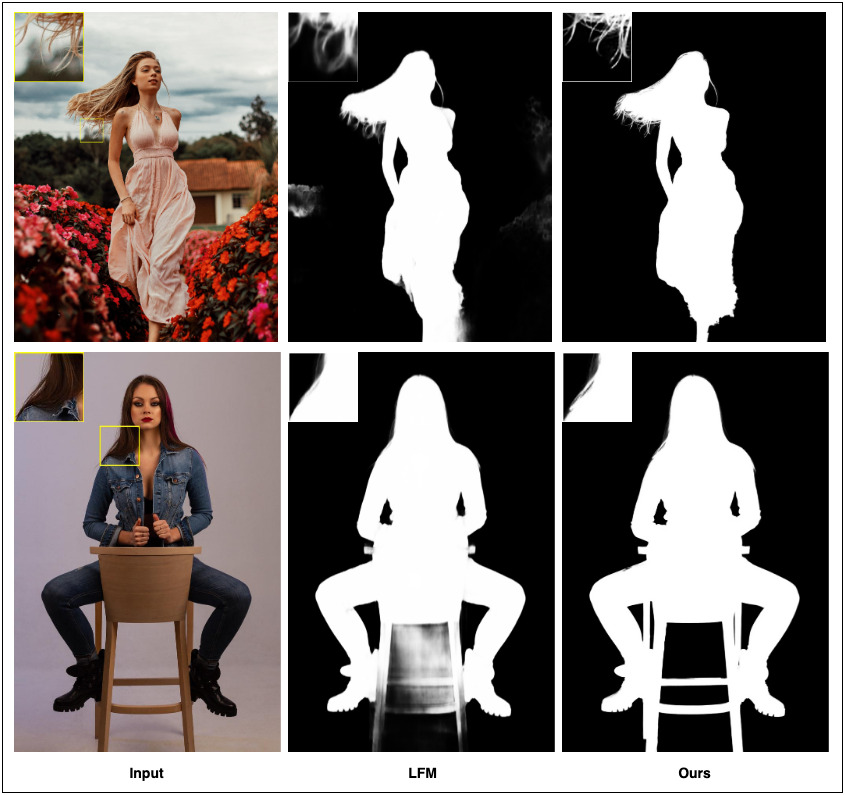}}{}
\makeatother 
\caption{{Alpha matte results by LFM and our SIM model directly from RGB input}}
\label{f-c49c208b3c08}
\end{figure}
\egroup
Deep learning has shown tremendous advancements in a number of classical computer vision areas. Xu et al.\unskip~\cite{1012980:21482488}  were the first to use deep learning to solve the matting problem. Their solution required a user generated trimap and the lower level semantics required to predict opacity was learned via a deep neural network. Several methods\unskip~\cite{1012980:21482475,1012980:21482493,1012980:21482474}  have been proposed hence which make use of a user trimap . A trimap contains the values of 0,255 and 128 which represent the absolute background, absolute foreground and the uncertain region in which the opacity must be predicted, respecitively. Recently, a promising line of work towards the aim of automatic matting has seen methods\unskip~\cite{1012980:21482496,1012980:21482499,1012980:21482495,1012980:21482489}  been proposed that go directly going from an RGB input to alpha matte. These methods attempt to solve the under constraint matting equation: 
\let\saveeqnno\theequation
\let\savefrac\frac
\def\dispfrac{\displaystyle\savefrac}
\begin{eqnarray}
\let\frac\dispfrac
\gdef\theequation{1}
\let\theHequation\theequation
\label{dfg-35ba06da5dd5}
\begin{array}{@{}l}\boldsymbol I=\boldsymbol\alpha\boldsymbol F+(1-\boldsymbol\alpha)\boldsymbol B\end{array}
\end{eqnarray}
\global\let\theequation\saveeqnno
\addtocounter{equation}{-1}\ignorespaces 
where I, $\alpha $, F and B stand for an image, the alpha map, the foreground image and background image respectively, without the help of a trimap. Without the trimap the network must estimate alpha across the whole image rather than only the unknown region which is a more challenging task. This task is more challenging as it requires knowledge of low level and higher level semantics. In non-automatic methods knowledge of high level semantics is carried by the trimap in absolute background and foreground regions.

Capturing the high level semantic features in an image consistently requires a model to be trained on a large quantity of data due to the large variations in image content. Also, in real-world use cases the foreground object can appear in a very different setting from the training data, for example a fashion model may appear with different accessories or in different poses. The foreground object may also not be pre-specified from before. In e-commerce for example, new products are constantly added to the catalogue. 

Previous automatic methods largely rely on expensive alpha annotations to learn such variations. Such annotations require skilled data annotators and are extremely expensive and time consuming to produce. Most current automatic methods either focus solely on humans or have trouble dealing with unseen object classes as well. Checking if an image is appropriate creates an additional layer of human involvement to either reject the image or create a trimap for it. Learning low level features on the other hand is more sample efficient can be further promoted by a better training set-up.

Motivated by these challenges, we propose a framework that can leverage cheap low quality annotations to learn robust semantic features and a fraction of high quality annotations used by other methods for learning of low-level features. Our simple yet powerful framework called \textbf{SIM}(\textbf{Salient Image Matting}) uses a novel saliency trimap network, inspired from salient object detection, that is able to produce trimaps of the most salient object in the image. The Salient Trimap Network(STN) is trained on trimaps generated from coarse annotations and a simple trimap generation scheme. This training allows the trimap network to accurately produce trimaps of  a large variety of foregrounds and be robust to large semantic variations in natural images. The output of the STN is then fed into a matting network for refinement of lower level semantics.  By decoupling the learning of these features we are able to provide guidance to the matting network of semantic information without a user generated trimap for arbitrary foreground objects. High quality data is used to finetune these models in an end-end fashion using the framework provided by Chen et al\unskip~\cite{1012980:21482499} . Further, for the task of image matting we propose a novel architecture with superior multi-scale feature representation than the common encoder-decoder architectures used for matting to learn the low level feature more efficiently.

\bgroup
\fixFloatSize{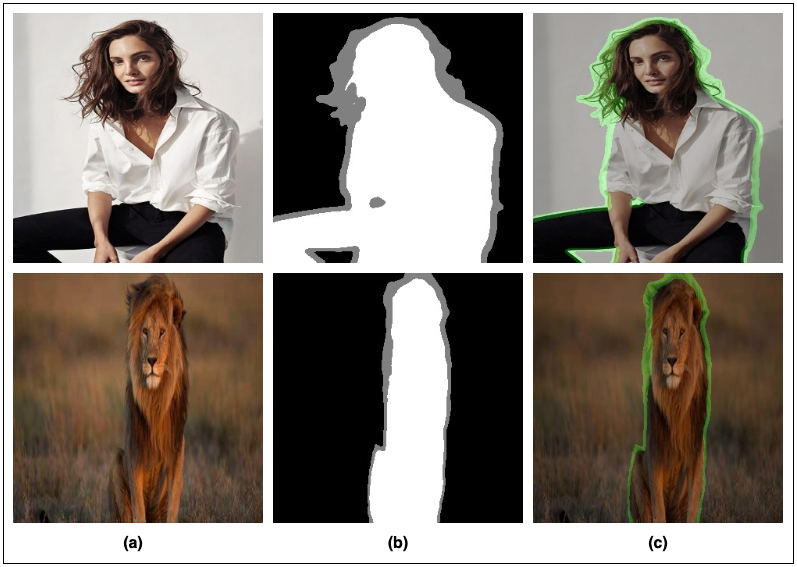}
\begin{figure}[!htbp]
\centering \makeatletter\IfFileExists{stn.png}{\includegraphics{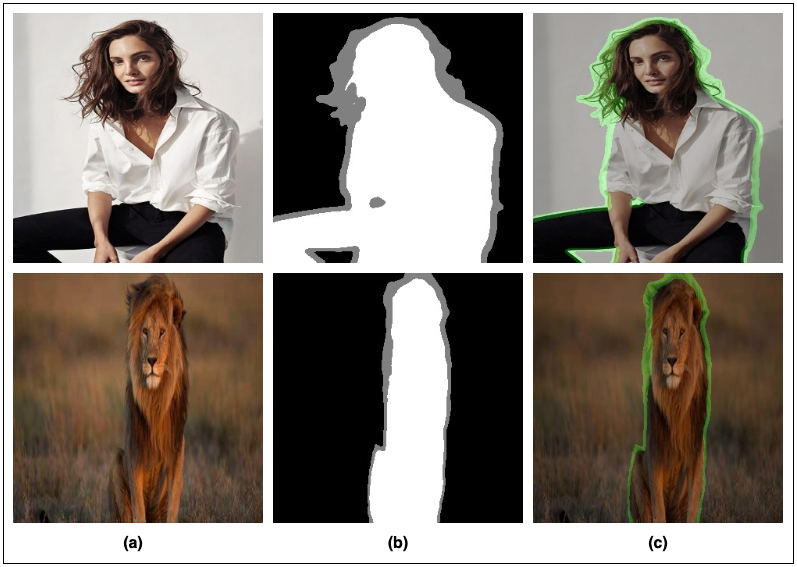}}{}
\makeatother 
\caption{{The intrinsic trimaps (b) produced by our STN. (a) are the input RGB images, (b) are the predicted trimaps and (c) are the trimap overlays on the input image.}}
\label{f-367df0ad7fd9}
\end{figure}
\egroup
Our framework can be viewed as an extension of existing matting models allowing them to handle large structural diversity and arbitrary foreground objects without user inputs. By using cheap coarse annotations we are able to achieve this aim in a very cost effective manner. We empirically evaluate the effectiveness of our method and achieve comparable results with the state-of-the-art interactive matting methods\unskip~\cite{1012980:21482488,1012980:21482491,1012980:21482474,1012980:21482475}  and outperform current automatic methods\unskip~\cite{1012980:21482496} . We also provide many examples of diverse images from the Internet to demonstrate our model's performance on real world use cases. 

We list our contributions as follows:

\begin{itemize}
  \item \relax To the best of our knowledge, we are the first to combine the tasks of object saliency detection with image matting to create an end to end framework which can produce high fidelity alpha mattes of arbitrary foreground objects in an image. This framework allows matting models to be used in more diverse settings that previously suggested . 
  \item \relax We propose a novel Salient Trimap Network which can generate trimaps for arbitrary foreground objects which can then be passed into a matting model for further refinement. We introduce a simple trimap generating scheme to effectively leverage coarsely annotated data for its pre-training. 
  \item \relax We introduce a novel architecture based on better multi-scale feature representation and fusion for the task of image matting that achieves superior performance to the common encoder-decoder architectures used for matting. 
\end{itemize}
  
\bgroup
\fixFloatSize{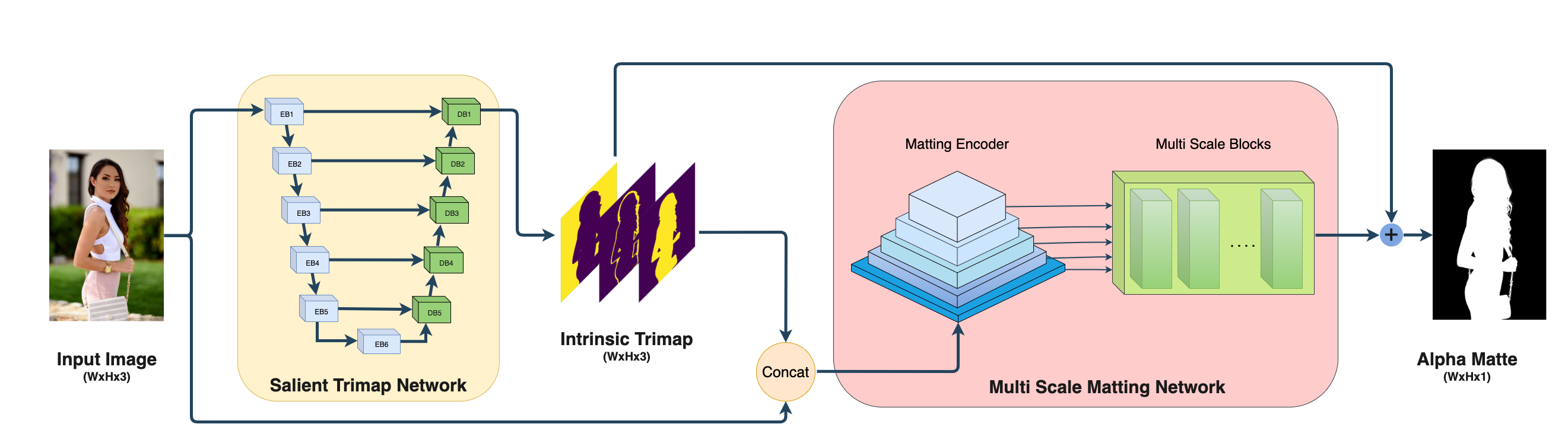}
\begin{figure*}[!htbp]
\centering \makeatletter\IfFileExists{drawio-1-page-1.png}{\includegraphics{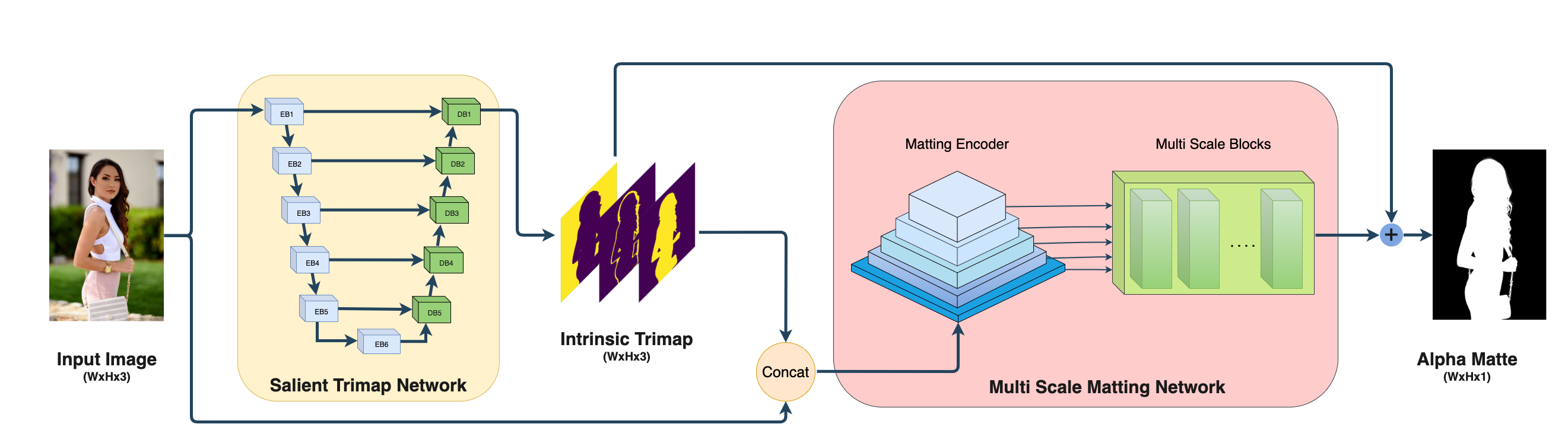}}{}
\makeatother 
\caption{{Overview of Salient Image Matting. An RGB image is provided as input, the STN generates an intrinsic trimap and the Multi Scale Matting network predicts the alpha matte. }}
\label{f-66a94a1ea035}
\end{figure*}
\egroup

\section{Related Work}
\textbf{Automatic Trimap Generation}: There have been a number of techniques that attempt to generate accurate trimaps for matting. \unskip~\cite{1012980:21482506,1012980:21482507,1012980:21482503} propose automatic trimap generation schemes that depends on depth information captured using special cameras. Hsieh et al.\unskip~\cite{1012980:21482501}  generated automatic trimaps from an input image along with its segmentation mask. Singh \& Jalal\unskip~\cite{1012980:21482508}  propose a method based on edge detection and thresholding to generate fine trimaps. Gupta \& Raman\unskip~\cite{1012980:21482504}  use super pixels and three saliency detection techniques to find the object of interest and then use a uniform erosion dilation scheme. Henry et al.\unskip~\cite{1012980:21482502}  use saliency detection to identify the foreground of an image, erode the boundaries and use lazy snapping and fuzzy c- means to identify the unknown pixels. \unskip~\cite{1012980:21482499}\unskip~\cite{1012980:21482471} used deep neural networks to generate trimaps.

\textbf{Image Matting: }Conventional image matting approaches include sampling based and affinity based methods. Sampling based approaches\unskip~\cite{1012980:21482462,1012980:21482461,1012980:21482463,1012980:21482464,1012980:21482467,1012980:21482468}  use color information of sampled pixels to estimate the alpha values of the unknown pixels in an image. Such methods build foreground and background color models to exploit natural image statistics. Affinity-based approaches\unskip~\cite{1012980:21482491},\unskip~\cite{1012980:21482455,1012980:21482450,1012980:21482457,1012980:21482492}  on the other hand propagates the alpha values of the known foreground and background regions to the unknown region. The predicted alpha map is determined by the affinity scores.

Recently deep learning has shown impressive results for matting. These methods adapt image segmentation encoder-decoder architectures to predict the alpha mattes from an RGB image input and an extra channel containing the user defined trimap. IndexNet Matting\unskip~\cite{1012980:21482493}  employs learnable index pooling and un-pooling for the encoder and decoder branch respectively. Context Aware Matting\unskip~\cite{1012980:21482474}  uses dual decoders to predict the alpha and foreground map. GCA Matting\unskip~\cite{1012980:21482475}  uses deep learning to exploit the user generated trimap by using a trimap guided attention mechanism.\unskip~\cite{1012980:21482448} tries to make matting work for a higher diversity of scenarios by taking in a background image as additional support for the matting network.\unskip~\cite{1012980:21482449} uses a discriminator to make the alpha matte for realistic.

\textbf{Automatic Matting  }Methods that only require a single RGB image input to generate a alpha matte provide great convenience in practical applications.  Automatic soft segmentation was introduced in\unskip~\cite{1012980:21482443}. \unskip~\cite{1012980:21482444,1012980:21482445} used segmentation masks as guidance for matting. SHM\unskip~\cite{1012980:21482499}  introduced a novel fusion method to fuse their trimap and matting prediction to make an end to end trainable network. HAttMatting\unskip~\cite{1012980:21482495}  tries to use attention mechanism in one network to train separate low and high level processing branches. SHM \unskip~\cite{1012980:21482499} and HAttMattin\unskip~\cite{1012980:21482495} introduced a fusion layer and hierarchical attention to carefully fuse low and high level feature. SHM also predict intermediate trimap while HAttMatting uses attention to highlight low level features in a different branch. LFM\unskip~\cite{1012980:21482496}  uses a dual decoder architecture to predict foreground and background and fuse via weighted fusion.\unskip~\cite{1012980:21482447}  makes use of coarse annotations to learn strong human semantics which are then supplied to their matting model after processing by a quality unification network. It does not produce a trimap but a coarse mask and a QN network for high and low quality data fusion. \unskip~\cite{1012980:21482446} used instance segmentation to identify objects in the image and then applied erosion dilation before matting. Their network was not end to end. 

\textbf{Multi-Scale Feature Representation}  Deep image matting architectures have conventionally been encoder- decoder based . Recent advancements in multi-scale feature representation and fusion are especially relevant for dense predication tasks like image matting. Unet\unskip~\cite{1012980:21482442}  and FPN\unskip~\cite{1012980:21482441}  were the first to introduce skip connection from an encoder to a decoder for fusion of low and high level features.  PANet\unskip~\cite{1012980:21482440}  added an extra bottom-up path aggregation pathway on top of FPN for further reconsideration and refinement of low level features. M2det\unskip~\cite{1012980:21482439}  tries to fuse multi-scale feature using a U-shape module. EfficientDet\unskip~\cite{1012980:21482438}  uses a repeating bi-pyramidal block with weighted attention for better muti-scale feature fusion and achieves high accuracy at a low compute budget. HRNet\unskip~\cite{1012980:21482436}  emphases the importance of high resolution feature processing by maintaining several parallel processing streams. \unskip~\cite{1012980:21482437}  provides for a nice survey of this area.
    
\section{Method}

\subsection{Overview}In order to capture high level and low level features separately we utilise two sub-networks in our SIM framework, a Salient Trimap Network (STN) and a matting network. This disentanglement allows us to use large amount of coarsely annotated data for training semantic features in the STN. The details on trimap generation from coarse data is described in section 3.2. The STN generates a three-channel output representing the background, unknown region and foreground respectively. Details about the STN are described in section 3.3. The matting network then takes the intrinsic trimap from the STN along with the original input and predicts a one channel alpha matte image. The outputs of the two sub-networks are then fused following the method of Chen et al. to produce the final alpha matte. Our SIM workflow can be seen inFigure~\ref{f-66a94a1ea035} . We also introduce a multi-scale layer, DensePN, that acts on a feature pyramid from an encoder. More details about our matting architecture are described in section 3.4. In the following subsections, we describe these in details.

\subsection{Adaptive Trimap Generation Scheme}In order to train our STN we require trimap ground truths, however these are expensive to produce. Instead, we collect coarsely annotated segmentation masks and then construct a scheme to best generate trimaps from such coarse masks. Coarsely annotated data can be collected from a variety of sources including PNGs or DeepLabV3+\unskip~\cite{1012980:21482428}  outputs of internet images. The choice of content in the coarse mask is important as it influences semantic knowledge of the STN, thus we source 20,856 images of humans and common objects, similar to those in the DUTS dataset.

\bgroup
\fixFloatSize{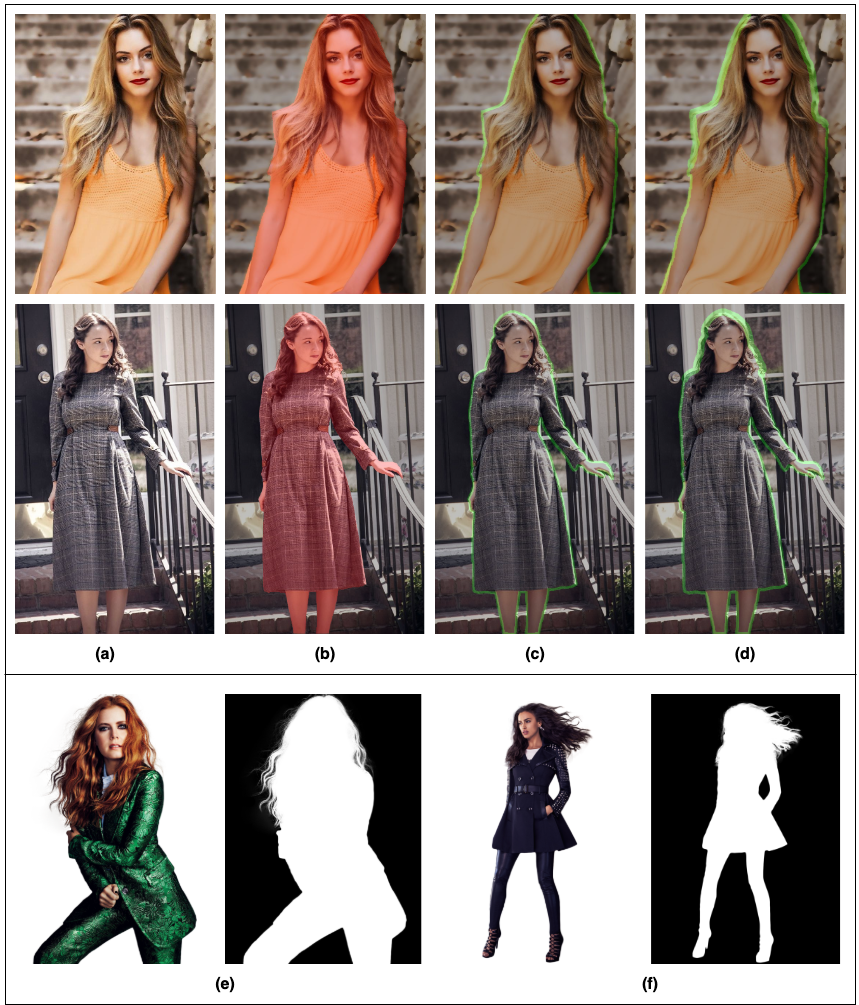}
\begin{figure}[t!]
\centering \makeatletter\IfFileExists{trimap-comparison.png}{\includegraphics{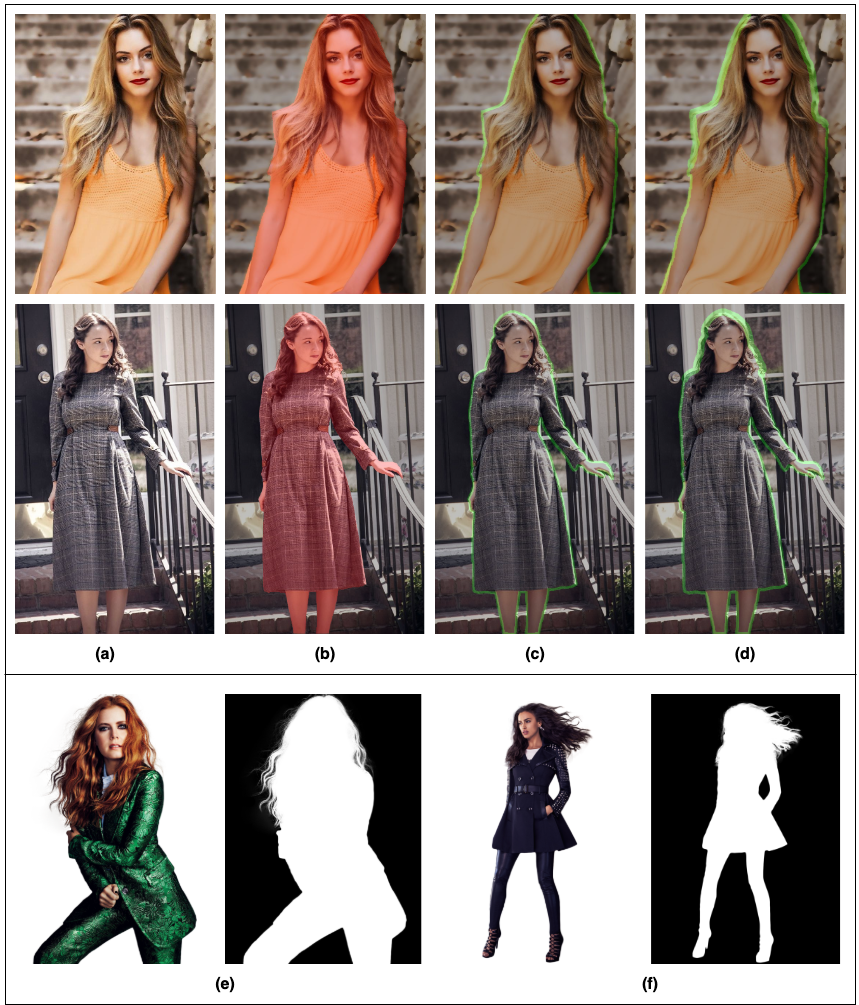}}{}
\makeatother 
\caption{{Examples of our trimap generation schemes. (a) input images, (b) Segmentation mask from DeepLabV3+ (c) Traditional rule based trimap with constant dilation everywhere, (d) Our trimap generation scheme which takes into account object size and features like hair. (e), (f) examples of PNG images from the internet to make trimaps for flying hair cases}}
\label{f-bb4724a477a1}
\end{figure}
\egroup
As our collection base is large, there is large variation in the size of images and the size of the foreground object. The colors of the foreground and background regions also resemble each other closely at times. As a result, common rule-based trimap generation schemes such as erosion-dilation and ones based on color information often results in inaccurate uncertainty regions. These methods also do not pay attention to common traits in coarse annotations. As show inFigure~\ref{f-bb4724a477a1} , we observe that coarse annotation around hair regions often under or over estimate foreground areas. For furry objects such as animals or soft toys we notice a similar pattern but less pronounced.  

We develop a simple but robust trimap generation scheme that takes into account the size of the object and object features such as hair and fur to generate trimaps from such coarse masks. We do so by classifying the boundary pixels of the coarse mask into three categories: hair, fur and solid and then dilating each separately. The classification is done as follows:

\begin{itemize}
  \item \relax For hair pixels, we apply a state of art human parsing network\unskip~\cite{1012980:21482435} on the image, to get mask of hair and body regions. The mask is then converted to only 2 classes - hair and non-hair. Each edge pixel in the segmentation mask is associated to the closest class in the parsed mask. As a result, edge pixels closer to the hair class of the parsed mask are classified as hair pixels. The remaining boundary pixels are marked as regular pixels. 
  \item \relax In images with animals or stuffed toys, all the boundary pixels are marked as fur pixels.  
  \item \relax If a pixel is neither detected as hair or fur then it is marked as a solid pixel. 
\end{itemize}
  To determine the rate of dilation of each pixel type in an image we define a metric  $D $ over the coarse image mask as a measure of the size of the salient object. The hair, fur and solid pixels are dilated by 3.5\%, 2.5\%, and 1.5\% of $D $ respectively. In our paper, we calculate $D $ as the maximum value of the object's distance map. Figure~\ref{f-bb4724a477a1}  shows our proposed adaptive trimap generation scheme as compared to the conventional erosion-dilation based method. Figure~\ref{f-bb4724a477a1} e,f also shows examples of internet PNG images along with their masks. While the opacity values in such images is not precise and often uncalibrated, accurate trimaps can yet be generated from these masks using standard erosion and dilation methods.

\subsection{Salient Trimap Network}Images of interest for alpha matting usually contain large semantic diversity. We cannot expect to learn these semantic features from matting data as it is expensive and usually limited. Instead of relying on an external input, we propose a Salient Trimap Network (STN) that predicts a trimap of the most salient foreground region. The STN's output is a 3 channel classification output that is a probabilistic estimate of the absolute background, unknown region and absolute foreground regions. The matting network need only to focus on learning low level details from high quality data. The STN can be based on any saliency object detection architecture; we choose to use an architecture based on U\ensuremath{^{2}}Net\unskip~\cite{1012980:21482473}  due to its ability to capture accurate semantics efficiently. The nested U-structure of U\ensuremath{^{2}}Net with the residual U-blocks enables the network to obtain multi-scale features without significantly degrading the feature map resolution, which helps STN to better classify the semantics among foreground, background and unknown region.
\bgroup
\fixFloatSize{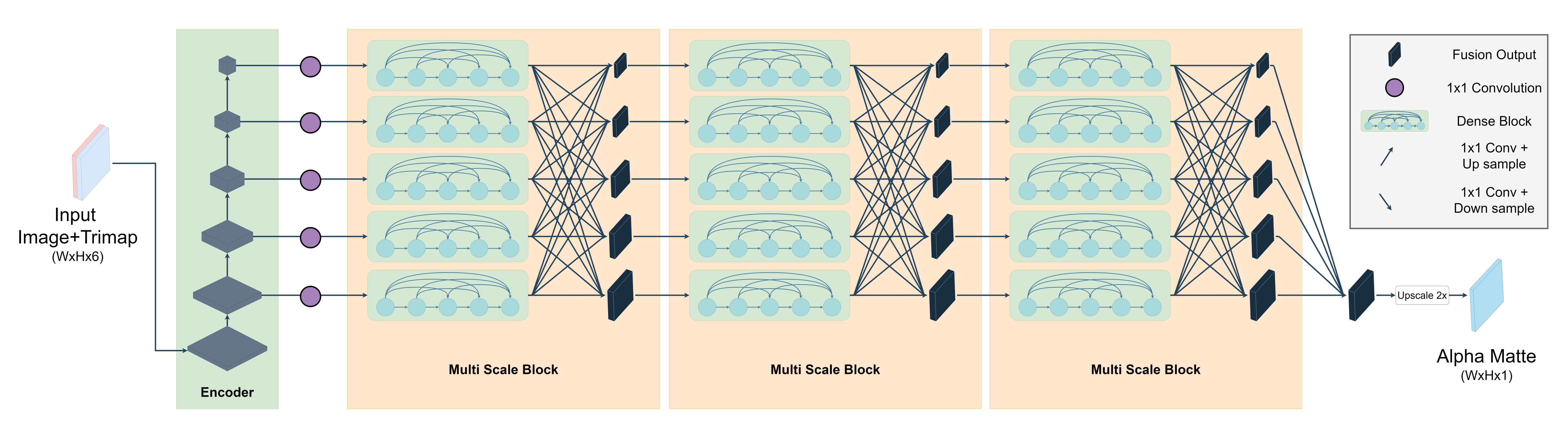}
\begin{figure*}[t!]
\centering \makeatletter\IfFileExists{mattingarch-page-3.png}{\includegraphics[width=\linewidth]{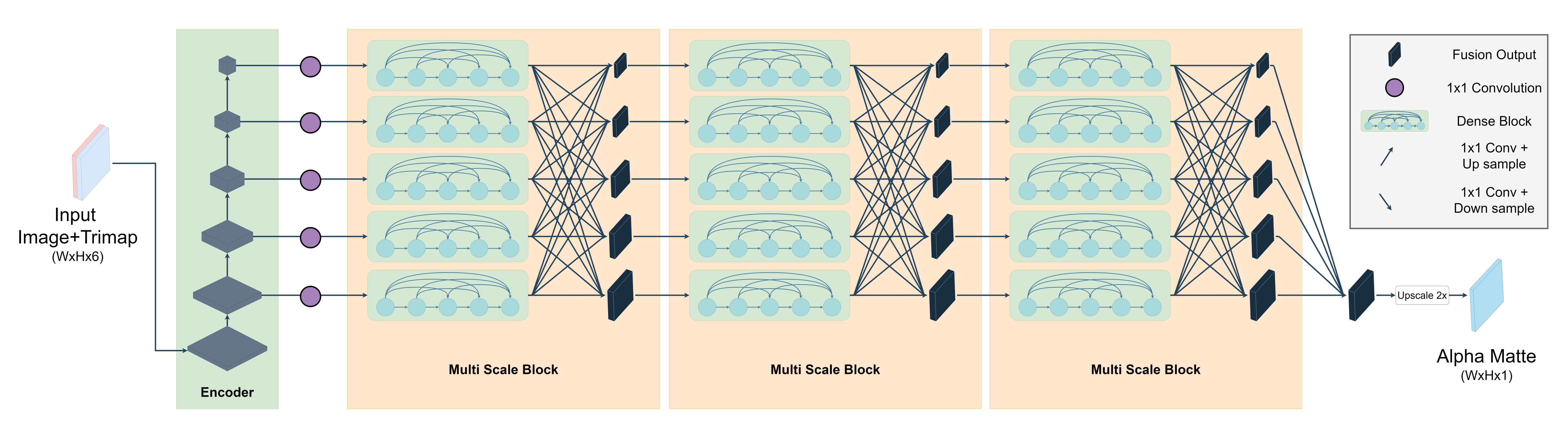}}{}
\makeatother 
\caption{{Architecture of the Multi Scale Fusion Model DensePN}}
\label{f-de0780dca78f}
\end{figure*}
\egroup

\subsection{Matting Network}\textbf{Network Design}  Encoder-decoder architectures only have one bottom up pathway which limits information flow from rich low level features and deep semantic features.  High resolution representations and multi-scale feature fusion are important for matting as they promote accurate localization and provide low level features with better context for opacity prediction. We crate a repeatable pyramid layer, called \textbf{DensePN}, which has parallel multi-resolution streams that are enriched with the other multi-resolution features. As shows in Figure~\ref{f-de0780dca78f}, each stream is a DenseBlock\unskip~\cite{1012980:21482433}  followed by fusion layer which brings all streams to the same resolution and performs a 1x1 convolution. The repeated convolution and fusion block allows for rich multi-scale feature enrichment at each resolution level. Finally, all stream are merged at the final prediction head to predict the alpha matte. We use ResNet34\unskip~\cite{1012980:21482432}  as our encoder. More deals of architecture can be found in supplementary materials. 

\textbf{Fusion}  The matting network produces an alpha matte with refined values only in the uncertain region suggested by the intrinsic trimap. As the matting network is trained only to refine low level details, it does not provide accurate values in the absolute foreground and background regions where semantic knowledge is necessary. The outputs of the STN and matting model are fused using the fusion technique proposed by Chen et al. Specifically if $F $,$B $ and $U $ denote the foreground,  background and unknown region probability maps predicted by the STN and $\alpha_m $ the output from the matting network then 
\begin{eqnarray*}\alpha_f=F + U*\alpha_m \end{eqnarray*}
This fusion allows for coarse semantics of the STN and fine details of the matting network to be combined effectively. Note that fusion is only required in the joint end-end training phase.

\subsection{Loss Functions}\textbf{STN Loss } The STN is trained for 3 way classification of the background, unknown region and foreground using the standard cross-entropy loss over each pixel. 

\textbf{Matting Loss } For the matting network pre-training, following\unskip~\cite{1012980:21482488}  we apply a combination of alpha prediction loss and composition loss. We also apply Laplacian Loss, proposed in\unskip~\cite{1012980:21482472}to further improve the performance of our network: 
\let\saveeqnno\theequation
\let\savefrac\frac
\def\dispfrac{\displaystyle\savefrac}
\begin{eqnarray}
\let\frac\dispfrac
\gdef\theequation{3}
\let\theHequation\theequation
\label{dfg-7bfc85e0c16a}
\begin{array}{@{}l}\mathcal{L}_{l a p}^{\alpha}=\sum_{i=1}^{5} 2^{i-1}\left\ensuremath{\Vert}L^{i}(\hat{\boldsymbol{\alpha}})-L^{i}(\boldsymbol{\alpha})\right\ensuremath{\Vert}_{1}\end{array}
\end{eqnarray}
\global\let\theequation\saveeqnno
\addtocounter{equation}{-1}\ignorespaces 
where $L^{i}(\boldsymbol{\alpha}) $is the $i $'th level of the laplacian pyramid of the alpha map. Our overall matting loss is defined as: 
\begin{eqnarray*}L_M = w_1\left\ensuremath{\Vert}\alpha_{p}-\alpha_{g}\right\ensuremath{\Vert}_{1}+w_2\left\ensuremath{\Vert}c_{p}-c_{g}\right\ensuremath{\Vert}_{1}+w_3L_{lap} \end{eqnarray*}
\textbf{Joint Loss }Theoretically, we could use the matting loss computed over the whole image for joint training as well, however we observed that STN forgets the semantic knowledge it learned from the coarse trimaps. To prevent this, we simultaneously pass in a batch of our trimap data along with the high quality matting data and apply a $L1 $ constraint over only the foreground regions of the coarse annotation. The joint loss is thus 
\begin{eqnarray*}\mathcal{L}_{F}=L_{M}+w_{4}\left\ensuremath{\Vert}\hat{1}\left(F_{s}>0\right) *\left(\hat{\mathbf{F}}_{\mathbf{S}}-\mathbf{F}_{\mathbf{S}}\right)\right\ensuremath{\Vert}_{1} \end{eqnarray*}
where F\ensuremath{_{S}} is the groundtruth foreground map and $\hat{1} $ is the indicator function. This soft $L1 $ constraint allows the joint network to optimise for delicate low and high level feature fusion between two models and also prevents the STN from forgetting its semantically rich features.

\subsection{Implementation Details}SIM is trained in three phases {\textemdash} STN pre-training, Matting Network pre-training and end-to-end training.

\subsubsection{STN Pre-training}The STN is pre-trained on trimaps generated from coarse segmentation masks generated by our proposed trimap generation scheme. In total, we have 20,856 coarse segmentation masks of which 8,919 are from the DUTS dataset\unskip~\cite{1012980:21482431}. These images include humans and animals in a variety of poses and many foreground objects such as those in DUTS dataset for the network to learn robust saliency detection. We also include trimaps generated by convention erosion-dilation for Adobe DIM data's\unskip~\cite{1012980:21482488}  alpha annotations so the network can jointly optimize for both sets during this stage. 

Following \unskip~\cite{1012980:21482473} , we apply deep supervision over all the six side outputs of the U\ensuremath{^{2}}Net with equal weight in this phase. Pre-trained weights of the U\ensuremath{^{2}}Net are used to initialize the network. The inputs are resized to 320x320 followed by random cropping of 288x288. Learning rate warmup and cosine annealing are used to train the network from a max learning rate of 8e-5 to 5e-6 with a batch size of 14.

\subsubsection{Matting Network Pre-training}The matting network is pre-trained with one hot encoded trimaps generated from standard erosion and dilation of the ground-truth alphas. Large kernels are randomly applied to make the matting network resistant to noisy trimaps. Following standard practice, random crops of 320x320,480x480 and 640x640 are taken centered around unknown regions and resized to 320x320. $L_M $ is only computed over the trimap region. All weights in $L_M $ are set to 1. Learning rate warmup and cosine annealing are used to train the network from a max learning rate of 1e-4 to 1e-5 with a batch size of 8.

\textbf{Data augmentation} Data augmentation is important for matting due to the limited number of unique high quality alphas. Flipping and color-jitter are applied randomly. Additionally, synthetic images composed between low quality COCO\unskip~\cite{1012980:21482430}  images and high quality matting foreground suffer from discrepancies in texture, noise, color quality and resolution in the foreground and background regions. The matting network can exploit these discrepancies and this can hinder its generalisability to real world images. To combat the domain gap of training on synthetic composed images, CAM\unskip~\cite{1012980:21482474}  used re-JPEGing and Gaussian blur and \unskip~\cite{1012980:21482448}  added gamma correction and random noise to background images before compositing. To circumvent the disparity in quality between foreground and background regions we create a dataset of HD backgrounds with non-salient foreground objects and use it for composition during training. Additionally, we also use Places365 Dataset\unskip~\cite{1012980:21482429}  to increase the texture and resolution diversity of the backgrounds.  

Additionally, for wild images the STN may classify certain thin structures within an object entirely into the unknown region like purse straps and chair legs. To better deal with such cases we add segmentation masks of solid object such as jewellery, purses and chairs obtained from internet images to the matting training data so the matting network learns to segment out such structures.

\subsubsection{End to End Training}For the joint training only high definition solid DIM images are used. The STN and Matting network are initialized with the best models in the pre-training stages. The output of the STN is resized to the original image size, to keep the matting net focused on low level features we continue the pre-training strategy for the matting network i.e random crops of 320x320,480x480 and 640x640 are taken and resized to 320x320, except the crop is now taken over the probabilistic trimap output from STN. Learning rate of 1e-5 and batch size of 4 is used to fit within the GPU memory. The batch-norm parameters of both networks are frozen in this stage due to the small batch size.
    
\section{Experiments}

\subsection{Datasets}For training of our matting architecture in the pre-training and end to end training phase, we use 227 training images of solid objects from the DIM dataset composed on backgrounds taken from COCO, Places356 and our HD background dataset equally to form a total training set of around 30,645 images. Results of our architecture with a trimap input are shown in Table~\ref{tw-d974969d6948} .

We compare our method against state of art trimap dependent methods on 1000 synthetic composed images made from compositing 50 unique foregrounds on 20 images from our HD dataset. Our test set contains 23 solid foreground images from the DIM dataset and 27 diverse high quality alpha mattes to evaluate the robustness of our framework. Our HD dataset does not contain any salient objects in the foreground as this may lead to ambiguity in detecting the most salient object by  the STN. 

Our  SIM model is trained and tested on a subset of DIM dataset excluding transparent and non-salient objects such as tree branches and wires. The 227 train images are composed on backgrounds from COCO, Places356 and our HD dataset not containing salient objects in their foreground. This is an important measure to prevent foreground ambiguity for the Salient Trimap Network.

\subsection{Results}We use the four commonly used metrics to evaluate the effectiveness of our proposed framework: Sum of Absolute Difference (SAD), Mean Square Error(MSE), Gradient Error and Connectivity Error as proposed in\unskip~\cite{1012980:21482487}.

\bgroup
\fixFloatSize{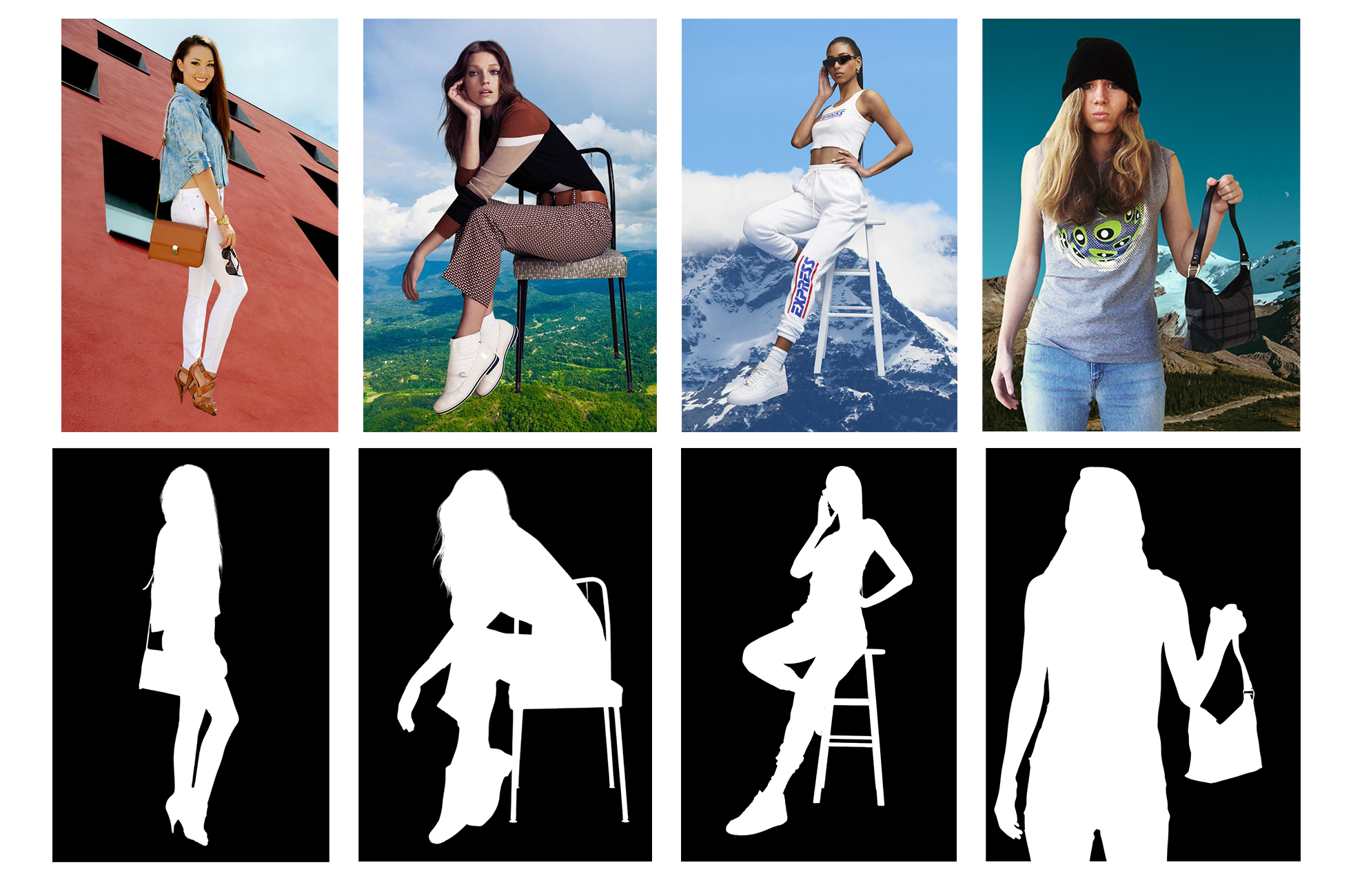}
\begin{figure}[t!]
\centering \makeatletter\IfFileExists{hdtest.png}{\includegraphics{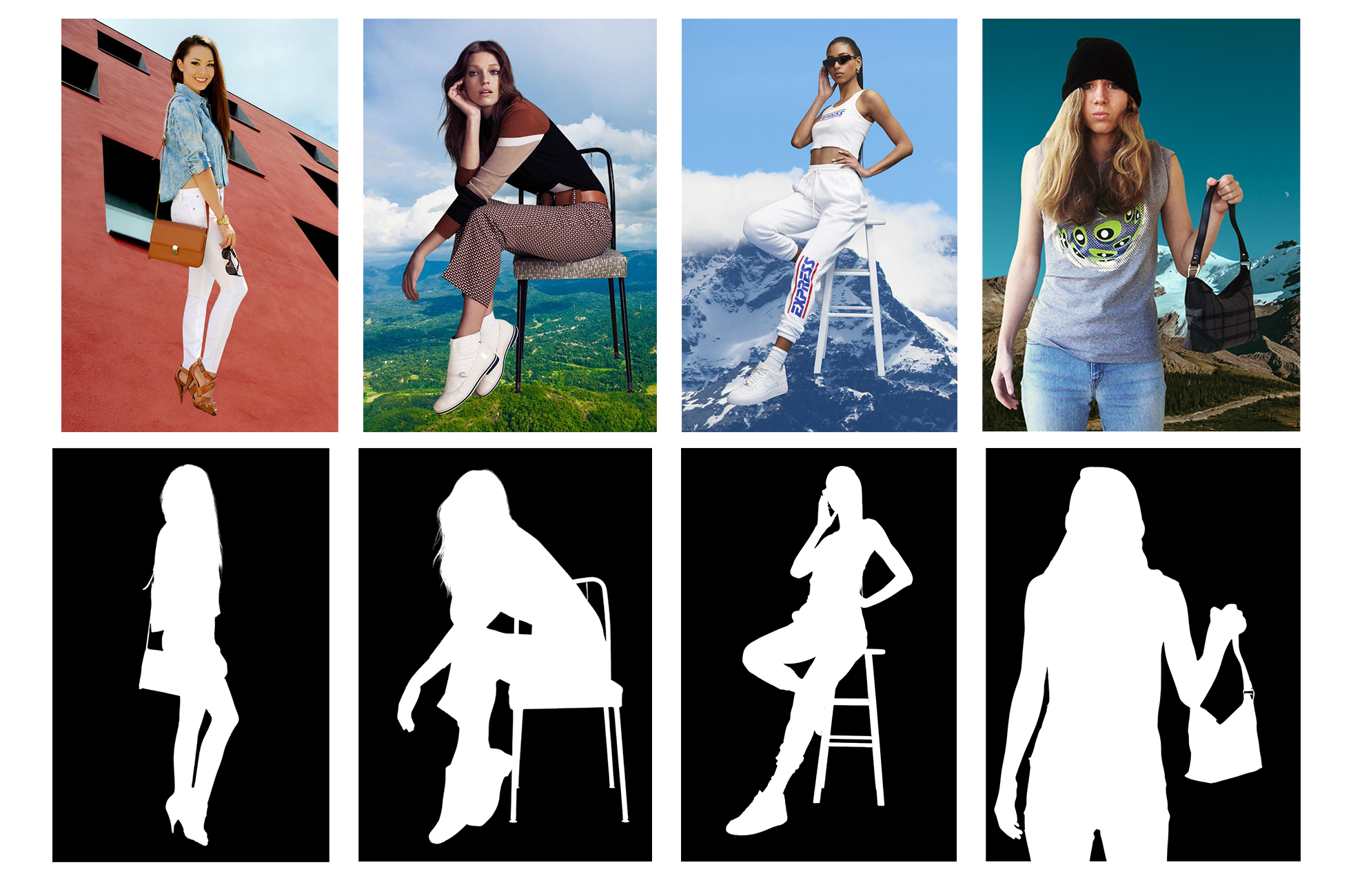}}{}
\makeatother 
\caption{{Samples from our extended test set along with their ground truth alpha matte.}}
\label{f-c131610b9c3f}
\end{figure}
\egroup

\bgroup
\fixFloatSize{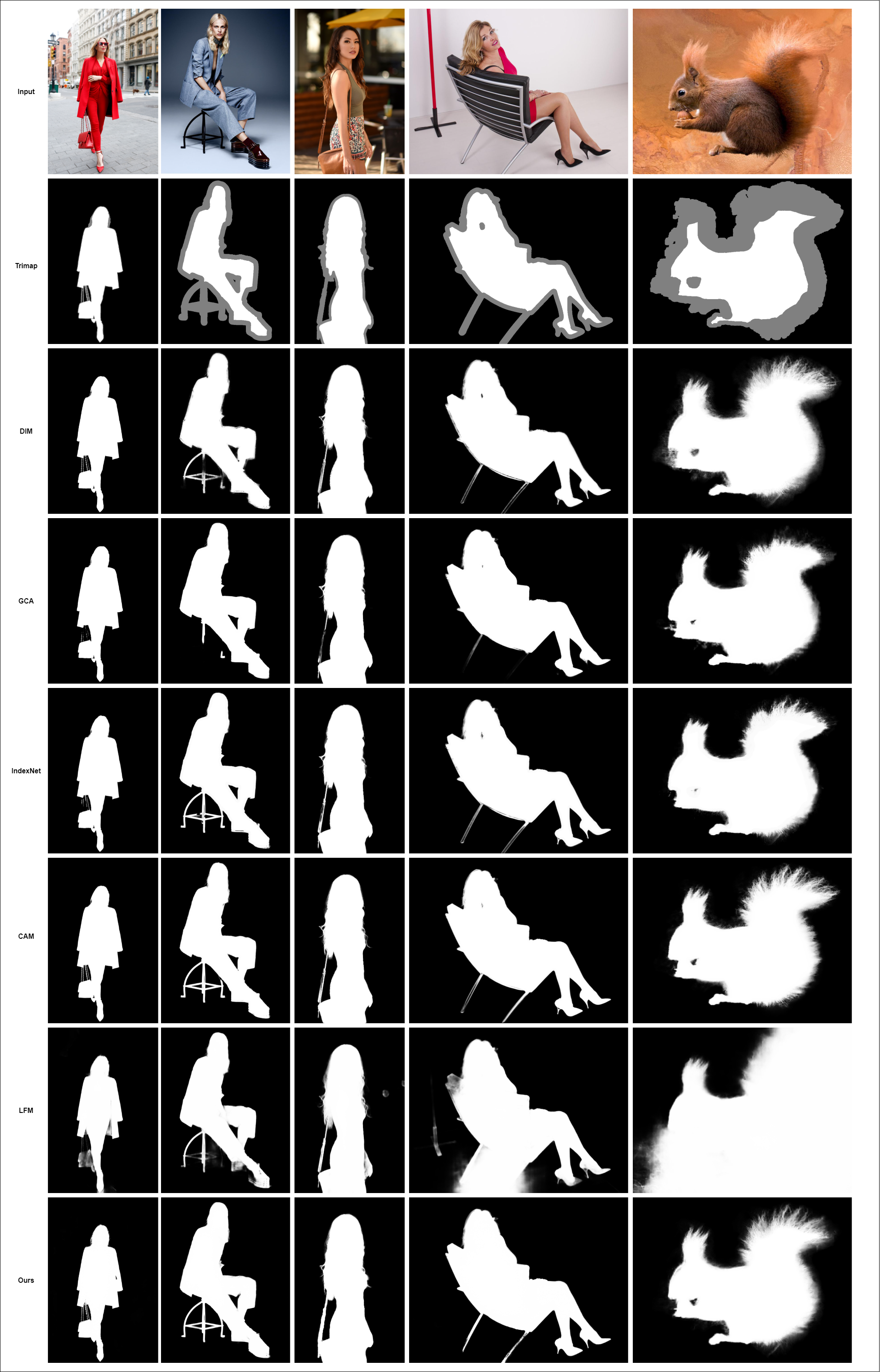}
\begin{figure*}[t!]
\centering \makeatletter\IfFileExists{results.png}{\includegraphics{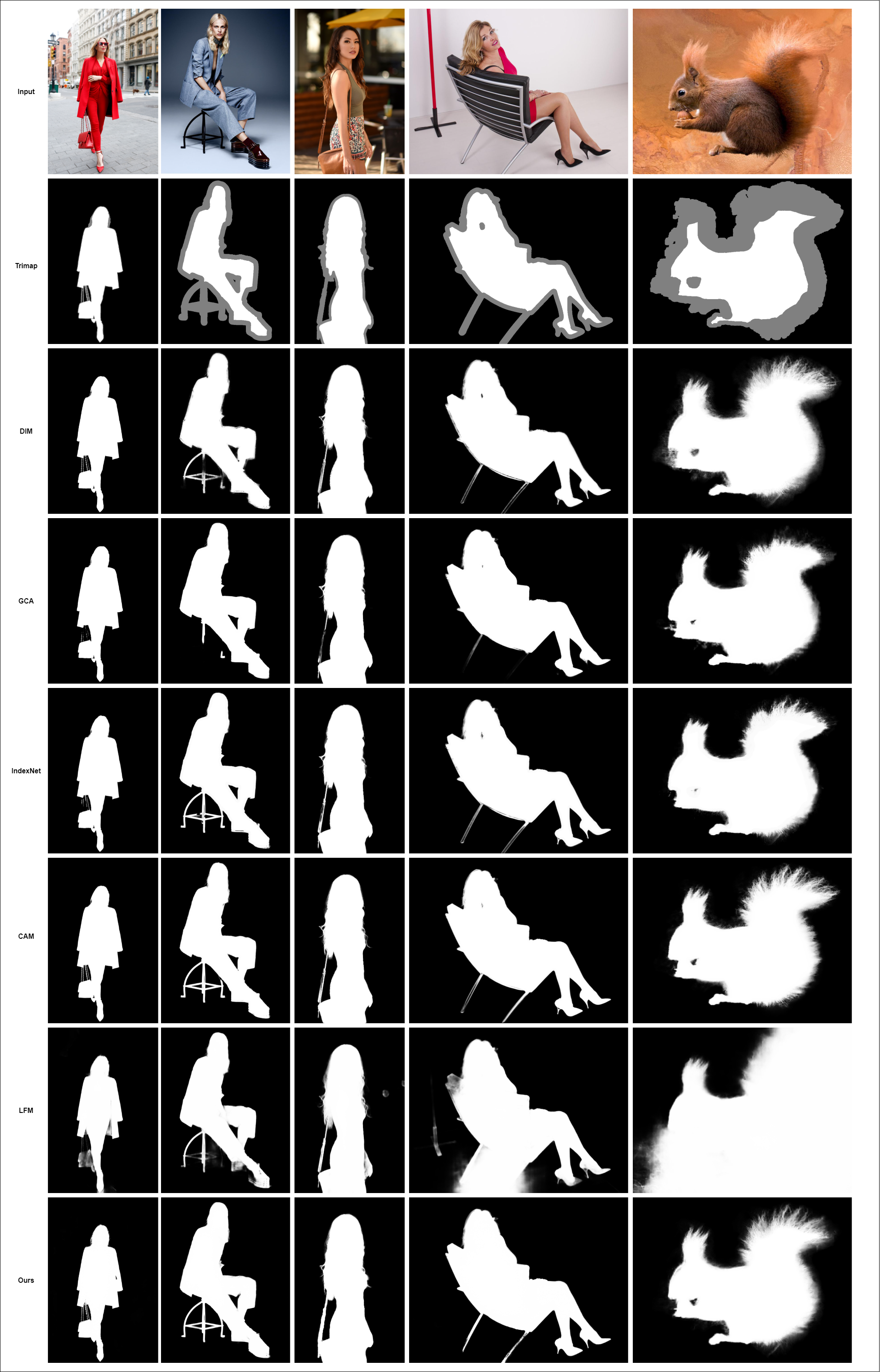}}{}
\makeatother 
\caption{{Comparison of our automatic method to other trimap dependent methods. LFM and Ours do not require a trimap.}}
\label{f-7662e2cd3ef6}
\end{figure*}
\egroup

\subsubsection{Trimap Based Matting Models}In the first phase of evaluation, we compare the effectiveness of our matting architecture using ground truth trimaps against six state of the art trimap based matting models. We compare our matting model to a few conventional and state-of-art deep learning methods.Table~\ref{tw-d974969d6948}  shows the results of this comparison. While KNN\unskip~\cite{1012980:21482491}, CFM\unskip~\cite{1012980:21482492}, DIM\unskip~\cite{1012980:21482488}, GCA\unskip~\cite{1012980:21482475} and IndexNet\unskip~\cite{1012980:21482493} predict only the alpha matte, Context Aware matting\unskip~\cite{1012980:21482474} simultaneously predicts the alpha matte as well as the foreground. In all cases except for DIM, we use the official public code repository. Since DIM does not have a publicly available model, we use a model trained by the authors of IndexNet as the accuracy closely matches what was reported in the paper.  Our network obtains superior performance as compared to other methods due to the ability of multi-scale feature fusion blocks to perceive fine-grained alpha mattes from adaptive semantics.
\begin{table}[!htbp]
\caption{{Comparison of Trimap Based Matting Models on test set} }
\label{tw-d974969d6948}
\centering 
\begin{threeparttable}

\def\arraystretch{1}
\ignorespaces 
\centering 
\begin{tabulary}{\linewidth}{p{\dimexpr.20\linewidth-2\tabcolsep}p{\dimexpr.20\linewidth-2\tabcolsep}p{\dimexpr.1667\linewidth-2\tabcolsep}p{\dimexpr.13569999999999999\linewidth-2\tabcolsep}p{\dimexpr.29759999999999998\linewidth-2\tabcolsep}}
\hline Model & SAD & MSE (x10\ensuremath{^{-2}}) & Gradient Error & Connectivity Error\\
\hline 
KNN &
  15.9 &
  3.4 &
  12.4 &
  17.8\\
CFM &
  12.3 &
  2.5 &
  9.5 &
  14.7\\
DIM &
  8.9 &
  1.7 &
  7.1 &
  9.9\\
IndexNet &
  7.8 &
  1.1 &
  6.3 &
  7.1\\
GCA &
  9.4 &
  2.3 &
  7.2 &
  13.5\\
CAM &
  7.9 &
  1.2 &
  6.5 &
  8.5\\\cline{1-1}\cline{2-2}\cline{3-3}\cline{4-4}\cline{5-5}
\cellcolor[HTML]{E5E5E5}{\textbf{Res34-DensePN(Ours)}} &
  \cellcolor[HTML]{E5E5E5}{\textbf{7.7}} &
  \cellcolor[HTML]{E5E5E5}{\textbf{1.1}} &
  \cellcolor[HTML]{E5E5E5}{\textbf{6.2}} &
  \cellcolor[HTML]{E5E5E5}{\textbf{7.9}}\\
\hline 
\end{tabulary}\par 
\begin{tablenotes}\footnotesize 
    
\item{All metrics computed are computed only over the transition region following the methods proposed in\unskip~\cite{1012980:21482488}}
\end{tablenotes}
\end{threeparttable}

\end{table}

\subsubsection{End to End Models}Of the few automatic methods, SHM and \unskip~\cite{1012980:21482447}  are designed exclusively for human images. HAttMatting\unskip~\cite{1012980:21482495}  and LFM are the algorithms that most closely match our use case. 

Since the training code of only LFM was available for us, we compare our end to end model on our test set. Table~\ref{tw-5d7e8a107dd0} compares our STN+DensePN without and with(STN) end to end finetuning against LFM.

\begin{table}[!htbp]
\caption{{Comparison of Automatic Matting Models on test set} }
\label{tw-5d7e8a107dd0}
\centering 
\begin{threeparttable}

\def\arraystretch{1}
\ignorespaces 
\centering 
\begin{tabulary}{\linewidth}{p{\dimexpr.2669\linewidth-2\tabcolsep}p{\dimexpr.14459999999999997\linewidth-2\tabcolsep}p{\dimexpr.2114\linewidth-2\tabcolsep}p{\dimexpr.1886\linewidth-2\tabcolsep}p{\dimexpr.1885\linewidth-2\tabcolsep}}
\hline Model & SAD & MSE (x10\ensuremath{^{-2}}) & Gradient Error & Connectivity Error\\
\hline 
LFM &
  69.9 &
  3.4 &
  28.9 &
  23.5\\\cline{1-1}\cline{2-2}\cline{3-3}\cline{4-4}\cline{5-5}
\cellcolor[HTML]{E5E5E5}{\textbf{STN+DensePN}} &
  \cellcolor[HTML]{E5E5E5}{\textbf{47.9}} &
  \cellcolor[HTML]{E5E5E5}{\textbf{2.0}} &
  \cellcolor[HTML]{E5E5E5}{\textbf{33.0}} &
  \cellcolor[HTML]{E5E5E5}{\textbf{34.2}}\\
\cellcolor[HTML]{E5E5E5}{\textbf{SIM (Ours)}} &
  \cellcolor[HTML]{E5E5E5}{\textbf{32.4}} &
  \cellcolor[HTML]{E5E5E5}{\textbf{1.5}} &
  \cellcolor[HTML]{E5E5E5}{\textbf{22.3}} &
  \cellcolor[HTML]{E5E5E5}{\textbf{35.4}}\\
\hline 
\end{tabulary}\par 
\begin{tablenotes}\footnotesize 
    
\item{All metrics computed are computed over the entire image}
\end{tablenotes}
\end{threeparttable}

\end{table}
Figure~\ref{f-7662e2cd3ef6}  shows the results on some images on trimap based and automatic methods. We see that LFM is not able to accurately capture object semantics due to relying on its limited alpha dataset while ours is able to identify and capture accurate semantics. We compare favorably to trimap based matting models without the need for user generated trimap. Our STN is able to accurately capture diverse semantics and pass on its guidance to our matting network for low level refinement. 
    
\section{Conclusion}
In this paper we propose an automatic, end to end model for salient image matting. The proposed model does not require any additional user inputs and produces high quality, alpha mattes for a variety of objects including humans, animals and more. Our testing shows demonstrates the robustness of our model on natural images. We achieve state-of-the-art results for automatic end to end matting and comparable results to other state of the art interactive methods which require externally provided trimaps along with the input image. This can be attributed to our two main sub networks {\textemdash} the STN which efficiently picks up the high level semantic details of the salient object in the image and the multi scale fusion matting network that learns the fine, low level features. We believe our framework will serve as a strong baseline for future automatic matting methods which are not restricted to just human or animal images but on a much wider range of objects.



%

\bibliographystyle{IEEEtran}

\bibliography{\jobname}

\begin{thebibliography}{10}
\providecommand{\url}[1]{#1}
\csname url@samestyle\endcsname
\providecommand{\newblock}{\relax}
\providecommand{\bibinfo}[2]{#2}
\providecommand{\BIBentrySTDinterwordspacing}{\spaceskip=0pt\relax}
\providecommand{\BIBentryALTinterwordstretchfactor}{4}
\providecommand{\BIBentryALTinterwordspacing}{\spaceskip=\fontdimen2\font plus
\BIBentryALTinterwordstretchfactor\fontdimen3\font minus
  \fontdimen4\font\relax}
\providecommand{\BIBforeignlanguage}[2]{{%
\expandafter\ifx\csname l@#1\endcsname\relax
\typeout{** WARNING: IEEEtran.bst: No hyphenation pattern has been}%
\typeout{** loaded for the language `#1'. Using the pattern for}%
\typeout{** the default language instead.}%
\else
\language=\csname l@#1\endcsname
\fi
#2}}
\providecommand{\BIBdecl}{\relax}
\BIBdecl

\bibitem{1012980:21482488}
N.~Xu, B.~Price, S.~Cohen, and T.~Huang, ``{Deep image matting},'' in
  \emph{{Proceedings of the IEEE Conference on Computer Vision and Pattern
  Recognition}}, 2017, pp. 2970--2979.

\bibitem{1012980:21482475}
Y.~Li and H.~Lu, ``{Natural Image Matting via Guided Contextual Attention},''
  \emph{{arXiv preprint arXiv:2001.04069}}, 2020.

\bibitem{1012980:21482493}
H.~Lu, Y.~Dai, C.~Shen, and S.~Xu, ``{Indices matter: Learning to index for
  deep image matting},'' in \emph{{Proceedings of the IEEE International
  Conference on Computer Vision}}, 2019, pp. 3266--3275.

\bibitem{1012980:21482474}
Q.~Hou and F.~Liu, ``{Context-aware image matting for simultaneous foreground
  and alpha estimation},'' in \emph{{Proceedings of the IEEE International
  Conference on Computer Vision}}, 2019, pp. 4130--4139.

\bibitem{1012980:21482496}
Y.~Zhang, L.~Gong, L.~Fan, P.~Ren, Q.~Huang, H.~Bao, and W.~Xu, ``{A late
  fusion cnn for digital matting},'' \emph{{Proceedings of the IEEE Conference
  on Computer Vision and Pattern Recognition}}, pp. 7469--7478, 2019.

\bibitem{1012980:21482499}
Q.~Chen, T.~Ge, Y.~Xu, Z.~Zhang, X.~Yang, and K.~Gai, ``{Semantic human
  matting},'' \emph{{Proceedings of the ACM International Conference on
  Multimedia}}, pp. 618--626, 2018.

\bibitem{1012980:21482495}
\BIBentryALTinterwordspacing
Y.~Qiao, Y.~Liu, X.~Yang, D.~Zhou, M.~Xu, Q.~Zhang, and X.~Wei,
  ``{Attention-Guided Hierarchical Structure Aggregation for Image Matting},''
  in \emph{{2020 IEEE/CVF Conference on Computer Vision and Pattern Recognition
  (CVPR)}}, 2020, pp. 13\,673--13\,682. [Online]. Available:
  \url{10.1109/CVPR42600.2020.01369}
\BIBentrySTDinterwordspacing

\bibitem{1012980:21482489}
J.~Liu, Y.~Yao, W.~Hou, M.~Cui, X.~Xie, C.~Zhang, and X.~sheng Hua, ``{Boosting
  Semantic Human Matting with Coarse Annotations},'' in \emph{{Proceedings of
  the IEEE/CVF Conference on Computer Vision and Pattern Recognition}}, 2020,
  pp. 8563--8572.

\bibitem{1012980:21482491}
Q.~Chen, D.~Li, and C.-K. Tang, ``{KNN matting},'' \emph{{IEEE transactions on
  pattern analysis and machine intelligence}}, vol.~35, no.~9, pp. 2175--2188,
  2013.

\bibitem{1012980:21482506}
O.~Wang, J.~Finger, Q.~Yang, J.~Davis, and R.~Yang, ``{Automatic Natural Video
  Matting with Depth},'' \emph{{15th Pacific Conference on Computer Graphics
  and Applications (PG'07}}, 2007.

\bibitem{1012980:21482507}
\BIBentryALTinterwordspacing
D.~Cho, S.~Kim, Y.-W. Tai, and I.~S. Kweon, ``{Automatic Trimap Generation and
  Consistent Matting for Light-Field Images},'' \emph{{IEEE Transactions on
  Pattern Analysis and Machine Intelligence}}, vol.~39, no.~8, pp. 1504--1517,
  2017. [Online]. Available: \url{10.1109/tpami.2016.2606397;
  https://dx.doi.org/10.1109/tpami.2016.2606397}
\BIBentrySTDinterwordspacing

\bibitem{1012980:21482503}
T.~Lu and S.~Li, ``{Image matting with color and depth information},''
  \emph{{Proceedings of the 21st International Conference on Pattern
  Recognition (ICPR2012)}}, 2012.

\bibitem{1012980:21482501}
M.~Hsieh and Lee, ``{Automatic trimap generation for digital image matting},''
  in \emph{{2013 Asia-Pacific Signal and Information Processing Association
  Annual Summit and Conference}}, 2013, pp. 1--5.

\bibitem{1012980:21482508}
\BIBentryALTinterwordspacing
S.~Singh and A.~S. Jalal, ``{Automatic generation of trimap for image
  matting},'' \emph{{International Journal of Machine Intelligence and Sensory
  Signal Processing}}, vol.~1, no.~3, pp. 232--232, 2014. [Online]. Available:
  \url{10.1504/ijmissp.2014.066425;
  https://dx.doi.org/10.1504/ijmissp.2014.066425}
\BIBentrySTDinterwordspacing

\bibitem{1012980:21482504}
V.~Gupta and S.~Raman, ``{Automatic trimap generation for image matting},''
  \emph{{2016 International Conference on Signal and Information Processing}},
  2016.

\bibitem{1012980:21482502}
\BIBentryALTinterwordspacing
C.~Henry and S.-W. Lee, ``{Automatic trimap generation and artifact reduction
  in alpha matte using unknown region detection},'' \emph{{Expert Systems with
  Applications}}, vol. 133, pp. 242--259, 2019. [Online]. Available:
  \url{10.1016/j.eswa.2019.05.019;
  https://dx.doi.org/10.1016/j.eswa.2019.05.019}
\BIBentrySTDinterwordspacing

\bibitem{1012980:21482471}
X.~Xiaoyongshen, Hongyungao, Chaozhou, and Jia, ``{Deep automatic portrait
  matting},'' \emph{{European Conference on Computer Vision}}, vol.~3, pp.
  6--6, 2016.

\bibitem{1012980:21482462}
A.~Karacan, E.~Erdem, and Erdem, ``{Image matting with kl-divergence based
  sparse sampling},'' \emph{{ICCV}}, 2015.

\bibitem{1012980:21482461}
C.~Rhemann and Rother, ``{A global sampling method for alpha matting},''
  \emph{{CVPR}}, 2011.

\bibitem{1012980:21482463}
Y.~Y. Chuang, B.~Curless, D.~H. Salesin, and R.~Szeliski, ``{A bayesian
  approach to digital matting},'' \emph{{CVPR}}, 2003.

\bibitem{1012980:21482464}
X.~Feng, X.~Liang, and Z.~Zhang, ``{A cluster sampling method for image matting
  via sparse coding},'' \emph{{ECCV}}, 2016.

\bibitem{1012980:21482467}
E.~Shahrian, D.~Rajan, B.~Price, and S.~Cohen, ``{Improving image matting using
  comprehensive sampling sets},'' \emph{{CVPR}}, 2013.

\bibitem{1012980:21482468}
J.~Wang and M.~F. Cohen, ``{Optimized color sampling for robust matting},''
  \emph{{CVPR}}, 2007.

\bibitem{1012980:21482455}
P.~Lee and Y.~Wu, ``{Nonlocal matting},'' \emph{{CVPR}}, 2011.

\bibitem{1012980:21482450}
Y.~Aksoy, M.~T.~O. Aydin, and Pollefeys, ``{Designing effective inter-pixel
  information flow for natural image matting},'' \emph{{CVPR}}, 2017.

\bibitem{1012980:21482457}
L.~Grady, T.~Schiwietz, A.~Shmuel, and R.~Westermann, ``{Random walks for
  interactive alpha-matting},'' \emph{{Proceedings of VIIP}}, 2005.

\bibitem{1012980:21482492}
A.~Levin, D.~Lischinski, and Y.~Weiss, ``{A closed-form solution to natural
  image matting},'' \emph{{IEEE transactions on pattern analysis and machine
  intelligence}}, vol.~30, no.~2, pp. 228--242, 2007.

\bibitem{1012980:21482448}
S.~Sengupta, V.~Jayaram, B.~Curless, S.~M. Seitz, and
  I.~Kemelmacher-Shlizerman, ``{Background Matting: The World is Your Green
  Screen},'' in \emph{{Proceedings of the IEEE/CVF Conference on Computer
  Vision and Pattern Recognition}}, 2020, pp. 2291--2300.

\bibitem{1012980:21482449}
S.~Lutz, K.~Amplianitis, and A.~Smolic, ``{AlphaGAN: Generative adversarial
  networks for natural image matting},'' \emph{{CoRR}}, vol. abs/1807.10088,
  2018.

\bibitem{1012980:21482443}
\BIBentryALTinterwordspacing
Y.~Aksoy, T.-H. Oh, S.~Paris, M.~Pollefeys, and W.~Matusik, ``{Semantic soft
  segmentation},'' \emph{{ACM Transactions on Graphics}}, vol.~37, no.~4, pp.
  1--13, 2018. [Online]. Available: \url{10.1145/3197517.3201275;
  https://dx.doi.org/10.1145/3197517.3201275}
\BIBentrySTDinterwordspacing

\bibitem{1012980:21482444}
X.~Shen, X.~Tao, H.~Gao, C.~Zhou, and J.-A. Jia, ``{Deep automatic portrait
  matting},'' \emph{{European Con- ference on Computer Vision}}, vol.~2, pp.
  3--3, 2016.

\bibitem{1012980:21482445}
B.~Zhu, Y.~Chen, J.~Wang, S.~Liu, B.~Zhang, and M.~Tang, ``{Fast deep matting
  for portrait anima- tion on mobile phone},'' \emph{{Proceedings of the 25th
  ACM inter- national conference on Multimedia}}, pp. 297--305, 2017.

\bibitem{1012980:21482447}
J.~Liu, Y.~Yao, W.~Hou, M.~Cui, X.~Xie, C.~Zhang, and X.-S. Hua, ``{Boosting
  Semantic Human Matting With Coarse Annotations},'' in \emph{{Proceedings of
  the IEEE/CVF Conference on Computer Vision and Pattern Recognition (CVPR)}},
  June 2020.

\bibitem{1012980:21482446}
G.~Hu and J.~J. Clark, ``{Instance Segmentation based Semantic Matting for
  Compositing Applications},'' \emph{{CoRR}}, vol. abs/1904.05457, 2019.

\bibitem{1012980:21482442}
O.~Ronneberger, P.~Fischer, and T.~Brox, ``{U-Net: Convolutional Networks for
  Biomedical Image Segmentation},'' \emph{{CoRR}}, vol. abs/1505.04597, 2015.

\bibitem{1012980:21482441}
T.-Y. Lin, P.~Doll\'{a}r, R.~B. Girshick, K.~He, B.~Hariharan, and S.~J.
  Belongie, ``{Feature Pyramid Networks for Object Detection},'' \emph{{CoRR}},
  vol. abs/1612.03144, 2016.

\bibitem{1012980:21482440}
S.~Liu, L.~Qi, H.~Qin, J.~Shi, and J.~Jia, ``{Path Aggregation Network for
  Instance Segmentation},'' \emph{{CoRR}}, vol. abs/1803.01534, 2018.

\bibitem{1012980:21482439}
Q.~Zhao, T.~Sheng, Y.~Wang, Z.~Tang, Y.~Chen, L.~Cai, and H.~Ling, ``{M2Det: A
  Single-Shot Object Detector based on Multi-Level Feature Pyramid Network},''
  \emph{{CoRR}}, vol. abs/1811.04533, 2018.

\bibitem{1012980:21482438}
M.~Tan, R.~Pang, and Q.~V. Le, ``{EfficientDet: Scalable and Efficient Object
  Detection},'' \emph{{CoRR}}, vol. abs/1911.09070, 2019.

\bibitem{1012980:21482436}
J.~Wang, K.~Sun, T.~Cheng, B.~Jiang, C.~Deng, Y.~Zhao, D.~Liu, Y.~Mu, M.~Tan,
  X.~Wang, W.~Liu, and B.~Xiao, ``{Deep High-Resolution Representation Learning
  for Visual Recognition},'' \emph{{CoRR}}, vol. abs/1908.07919, 2019.

\bibitem{1012980:21482437}
L.~Liu, W.~Ouyang, X.~Wang, P.~W. Fieguth, J.~Chen, X.~Liu, and
  M.~Pietik{\"{a}}inen, ``{Deep Learning for Generic Object Detection: A
  Survey},'' \emph{{CoRR}}, vol. abs/1809.02165, 2018.

\bibitem{1012980:21482428}
L.-C. Chen, Y.~Zhu, G.~Papandreou, F.~Schroff, and H.~Adam, ``{Encoder-Decoder
  with Atrous Separable Convolution for Semantic Image Segmentation},''
  \emph{{CoRR}}, vol. abs/1802.02611, 2018.

\bibitem{1012980:21482435}
P.~Li, Y.~Xu, Y.~Wei, and Y.~Yang, ``{Self-Correction for Human Parsing},''
  \emph{{CoRR}}, vol. abs/1910.09777, 2019.

\bibitem{1012980:21482473}
X.~Qin, Z.~Zhang, C.~Huang, M.~Dehghan, O.~R. Zaiane, and M.~Jagersand,
  ``{U2-Net: Going deeper with nested U-structure for salient object
  detection},'' \emph{{Pattern Recognition}}, vol. 106, p. 107404, 2020.

\bibitem{1012980:21482433}
G.~Huang, Z.~Liu, and K.~Q. Weinberger, ``{Densely Connected Convolutional
  Networks},'' \emph{{CoRR}}, vol. abs/1608.06993, 2016.

\bibitem{1012980:21482432}
K.~He, X.~Zhang, S.~Ren, and J.~Sun, ``{Deep Residual Learning for Image
  Recognition},'' \emph{{CoRR}}, vol. abs/1512.03385, 2015.

\bibitem{1012980:21482472}
S.~Niklaus and F.~Liu, ``{Context-aware synthesis for video frame
  interpolation},'' in \emph{{Proceedings of the IEEE Conference on Computer
  Vision and Pattern Recognition}}, 2018, pp. 1701--1710.

\bibitem{1012980:21482431}
\BIBentryALTinterwordspacing
L.~Wang, H.~Lu, Y.~Wang, M.~Feng, D.~Wang, B.~Yin, and X.~Ruan, ``{Learning to
  Detect Salient Objects with Image-Level Supervision},'' in \emph{{2017 IEEE
  Conference on Computer Vision and Pattern Recognition (CVPR)}}, 2017, pp.
  3796--3805. [Online]. Available: \url{10.1109/CVPR.2017.404}
\BIBentrySTDinterwordspacing

\bibitem{1012980:21482430}
T.-Y. Lin, M.~Maire, S.~J. Belongie, L.~D. Bourdev, R.~B. Girshick, J.~Hays,
  P.~Perona, D.~Ramanan, P.~Doll\'{a}r, and C.~L. Zitnick, ``{Microsoft COCO:
  Common Objects in Context},'' \emph{{CoRR}}, vol. abs/1405.0312, 2014.

\bibitem{1012980:21482429}
B.~Zhou, A.~Lapedriza, A.~Khosla, A.~Oliva, and A.~Torralba, ``{Places: A 10
  million Image Database for Scene Recognition},'' \emph{{IEEE Transactions on
  Pattern Analysis and Machine Intelligence}}, 2017.

\bibitem{1012980:21482487}
\BIBentryALTinterwordspacing
C.~Rhemann, C.~Rother, J.~Wang, M.~Gelautz, P.~Kohli, and P.~Rott, ``{A
  perceptually motivated online benchmark for image matting},'' in \emph{{2009
  IEEE Conference on Computer Vision and Pattern Recognition}}, 2009, pp.
  1826--1833. [Online]. Available: \url{10.1109/CVPR.2009.5206503}
\BIBentrySTDinterwordspacing

\end{thebibliography}
\vfill
\end{document}